\documentclass[conference]{IEEEtran}
\IEEEoverridecommandlockouts
\usepackage{cite}
\usepackage{amsmath,amssymb,amsfonts}
\usepackage{algorithmic}
\usepackage{graphicx}
\usepackage{textcomp}
\usepackage{xcolor}
\usepackage{times}
\usepackage{soul}
\usepackage{url}
\usepackage[utf8]{inputenc}
\usepackage[small]{caption}
\usepackage{graphicx}
\usepackage{amsmath}
\usepackage{booktabs}
\usepackage{booktabs}
\usepackage{bm}
\usepackage{url}
\usepackage{mathrsfs}
\usepackage{multirow}
\usepackage{balance}
\usepackage{amsthm}
\usepackage{amsmath}
\usepackage{amstext}
\usepackage{amssymb}
\usepackage[ruled,vlined,linesnumbered]{algorithm2e}
\usepackage{subcaption}
\usepackage{overpic}
\usepackage{color}
\usepackage{mathbbol}
\usepackage{subcaption}
\usepackage{enumitem}
\usepackage{amsmath}
\newcommand{\RNum}[1]{\uppercase\expandafter{\romannumeral #1\relax}}
\def\BibTeX{{\rm B\kern-.05em{\sc i\kern-.025em b}\kern-.08em
    T\kern-.1667em\lower.7ex\hbox{E}\kern-.125emX}}
\begin{document}

\title{Learning Credible Deep Neural Networks with Rationale Regularization}


\author{\IEEEauthorblockN{Mengnan Du, Ninghao Liu, Fan Yang, Xia Hu}
\IEEEauthorblockA{Department of Computer Science and Engineering, Texas A\&M University \\
\{dumengnan, nhliu43, nacoyang, xiahu\}@tamu.edu}}

\maketitle

\begin{abstract}
Recent explainability related studies have shown that state-of-the-art DNNs do not always adopt correct evidences to make decisions. It not only hampers their generalization but also makes them less likely to be trusted by end-users. In pursuit of developing more credible DNNs, in this paper we propose CREX, which encourages DNN models to focus more on evidences that actually matter for the task at hand, and to avoid overfitting to data-dependent bias and artifacts. Specifically, CREX regularizes the training process of DNNs with rationales, \emph{i.e.}, a subset of features highlighted by domain experts as justifications for predictions, to enforce DNNs to generate local explanations that conform with expert rationales. Even when rationales are not available, CREX still could be useful by requiring the generated explanations to be sparse. Experimental results on two text classification datasets demonstrate the increased credibility of DNNs trained with CREX. Comprehensive analysis further shows that while CREX does not always improve prediction accuracy on the held-out test set, it significantly increases DNN accuracy on new and previously unseen data beyond test set, highlighting the advantage of the increased credibility.
\end{abstract}


\maketitle

\section{Introduction}
There has been an increasing interest recently in developing explainable deep neural networks (DNNs)~\cite{montavon2018methods,du2018techniques,du2018towards,du2019attribution}. To this end, a DNN model should be able to provide intuitive explanations for its predictions. 
Explainability could shed light into the decision making process of DNNs and thus increase their acceptance by end-users.
However, explainability alone is insufficient for DNNs to be \textit{credible}~\cite{wang2018learning}, unless the provided explanations conform with the well-established domain knowledge. That is to say, correct evidences should be adopted by the networks to make predictions. The incredibility issue has been observed in various DNN systems.
For instance, in question answering (QA) tasks, DNNs rely more on function words rather than pay attention to task-specific verbs, nouns and adjectives to make decisions~\cite{mudrakarta2018did,rychalska2018does}. Similarly, in image classification, CNNs may make decisions solely according to background within images, rather than paying attention to evidences relevant to the objects of interest~\cite{ribeiro2016should}. 

In this work, we define credible DNNs as the models that could provide explanations to their predictions, while at the same time the explanations are consistent with the well-established domain knowledge. Considering that correct evidences are employed in decision making process, it would be easier for credible DNNs to build up trust among practitioners and end-users.
In addition, credible DNNs could have better generalization capability comparing to untrustable ones. Since credible DNNs have truly grasped useful knowledge instead of memorizing unreliable  \emph{dataset-specific biases} and \emph{artifacts}, they could maintain high prediction accuracy for those unseen data instances beyond the training dataset.

It is possible to enhance the credibility and generalization of DNNs from two perspectives: dataset and model training. The former category tackles this problem by constructing datasets with larger quantity and higher quality.
Any training data may contain some biases, either intrinsic noise or additional signals inadvertently introduced by human annotators~\cite{gururangan2018annotation}. DNNs not only rely on these biases to make decisions, but also could amplify them~\cite{bolukbasi2016man}, which partly leads to the low credibility and low generalization problem. Some work has developed debiased datasets either by filtering out bias data, or constructing new datasets in an adversarial manner~\cite{zellers2018swag}.
Nevertheless, this scheme cannot fully eliminate bias, which still could affect model performance.
The second category aims at regulating the training of DNNs using domain knowledge established by humans. 
This is motivated by the observation that purely data-driven learning could lead to counter-intuitive results~\cite{hu2016harnessing}. Thus it is desirable to combine DNNs with the domain knowledge that humans utilize to understand the world, which has been proven beneficial in a lot of learning problems~\cite{mihaylov2018knowledgeable,hu2016harnessing,zhang2016rationale}. Therefore, we follow the second strategy using domain knowledge to enhance the credibility of DNNs.

Nevertheless, regulating the training of DNNs with domain knowledge to promote model credibility is still a technically challenging problem. First, one difficulty lies in how to accurately obtain and effectively utilize DNNs' attention towards input features. 
Although DNN local explanations could identify the contributions of each input feature towards a specific model prediction~\cite{ribeiro2016should}, it is still challenging to incorporate explanation into the end-to-end back-propagation procedure to influence model parameter update.
The second challenge is how to use domain knowledge to regularize the models' attention and force models to focus on correct evidences. Previous work have demonstrated that domain knowledge is beneficial in terms of promoting prediction accuracy of DNNs. For instance, structured knowledge in the form of logical rules can be transferred to the weights of DNNs through iterative distillation process~\cite{hu2016harnessing}. However, it is still unclear how to utilize knowledge to guide the attention of a DNN. 

\begin{figure}
  \centering
  \includegraphics[width=0.95\linewidth]{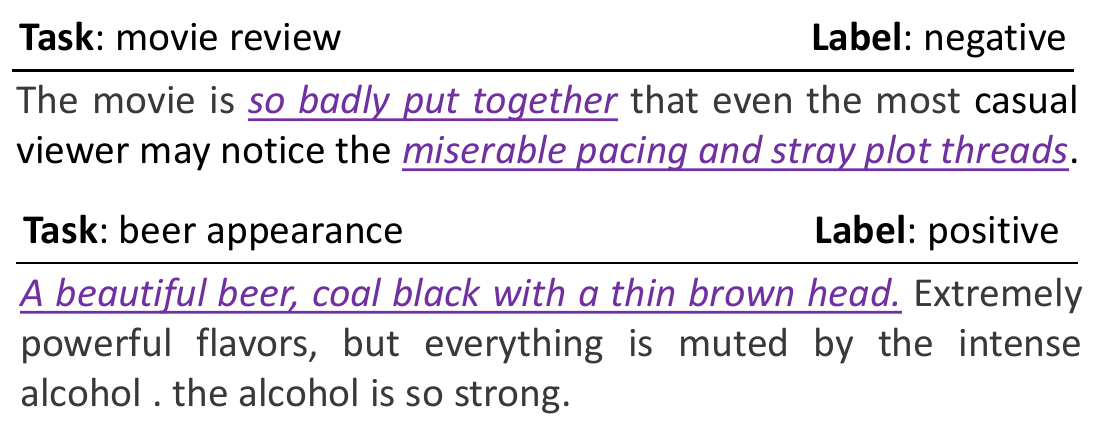}
  \caption{Two examples of expert rationale: words marked with purple color, for movie review and product review respectively. }
  \label{fig:rationale-example}
\end{figure}

To overcome the above challenges, we propose to explore whether a specific kind of domain knowledge, called \emph{rationale}, would be useful in terms of enhancing DNN credibility. A rationale is a subset of features highlighted by annotators and regarded to be more important in predicting an instance~\cite{zaidan2007using,lei2016rationalizing}, with illustrative examples shown in Fig.~\ref{fig:rationale-example}. The rationales are utilized to direct the model's attention, enabling it to tease apart useful evidence from noises and pushing it to pay more attention to relevant features. Rationales have been applied to the training process of SVMs~\cite{zaidan2007using,donahue2011annotator} to enhance predictive performance. Another benefit of rationales is that they require little effort to obtain~\cite{mcdonnell2016relevant}, thus they are possible to be widely applied in different applications.  

In this work, we propose CREX (\underline{CR}edible \underline{EX}planation), an approach regularizing DNNs to utilize correct evidences to make decisions, in order to promote their credibility and generalization capability. 
The intuition behind CREX is to use external knowledge to regulate the DNN training process. For those training instances coupled with expert rationales, we require the DNN model to generate local explanations that conform with the rationales. Even when expert rationales are not available, CREX can still promote model performance by requiring the generated explanations to be sparse. 
Through experiments on text classification tasks, we demonstrate that our trained DNNs generally rely on correct evidences to make predictions. Besides, our trained DNNs
generalize much better on new and previously unseen inputs beyond test set.
The major contributions of this paper are summarized as follows:
\begin{itemize}[leftmargin=*]
\setlength\itemsep{0em}
\item We propose a method to regularize the training of DNNs, called CREX, which aims to enable trained DNNs to focus on correct evidences to make decisions.
\item CREX is widely applicable to different variants of DNNs. We demonstrate its applicability via three standard architectures, including CNN, LSTM and self-attention model.
\item Experimental results on two text classification datasets validate that our trained DNNs could generate explanations aligning well with expert rationales and show good generalization properties on data beyond test set.
\end{itemize}

\section{Related Work}
In this section, we briefly present reviews for several research areas closely relevant to our work.

\subsection{DNN Interpretability}
DNNs are often regarded as black-boxes and criticized by the lack of interpretability. Towards this end, there is a wide range of work targeting to derive explanations and shed some insights into the decision making process of DNNs~\cite{doshi2017towards,du2018techniques}.
These work can be grouped into two main categories: global and local explanation,
depending on whether the goal is to understand how the DNN works globally or how DNN makes a specific prediction~\cite{montavon2018methods}. Most current work focus on augmenting DNNs with interpretability~\cite{ribeiro2016should,yang2019evaluating,liu2019representation}, while employing explanation to enhance the performance of DNN models has seldom been explored. In this work, we aim to take advantage of DNN local explanation to promote the generalization performance of DNN classifiers.

\subsection{Model Credibility and Generalization}
Despite the high performance of DNN models on test set,
recent work shows that these models heavily rely on dataset bias instead of true evidences to make decisions~\cite{agrawal2018don}. For instance, a DNN local explanation approach analyzes three question
answering models, showing that these models often ignore important part of the questions, e.g., verbs in questions carry little influence for the DNN decisions, and rely on irrelevant words to make decisions~\cite{mudrakarta2018did}. Similarly, for binary husky and wolf classification task, the CNN simply makes decisions according to whether there is snow within an image or not, rather than pays attention to evidences relevant to animals~\cite{ribeiro2016should}. This makes the DNN models unreliable and hampers their generalization. In addition, this also makes these models fragile and easily broken by adversarial samples.

\subsection{Unwanted Dataset Bias}
Datasets may contain lots of unwanted bias and artifacts, either explicit ones, e.g., gender and ethnic biases, or implicit ones. DNNs not only rely on these biases to make decisions, but also could amplify them~\cite{bolukbasi2016man}, which partly lead to the low credibility and low generalization of DNNs on unseen data. In order to alleviate the influence of unwanted dataset bias to models' performance, one line of work tackles this problem by regulating the training of models~\cite{hendricks2018women,ramakrishnan2018overcoming}, while some others consider to construct more challenging datasets by eliminating biases and annotation artifacts~\cite{zellers2018swag,rajpurkar2018know}.

\subsection{Combining Human Knowledge with DNNs} 
Some work enhances DNN models with human-like common sense
to make them more credible and robust.
For instance, the attention of RNN is regularized with human attention values derived from eye-tracking corpora~\cite{barrett2018sequence}. Structured knowledge such as logical rules are transfered to the weight of DNNs through iterative distillation process~\cite{hu2016harnessing}. Besides, rationales are augmented to the training process of CNN models~\cite{zhang2016rationale}, linear classification model~\cite{wang2018learning}, and SVMs~\cite{donahue2011annotator}. These work indicates that human knowledge has indeed promoted the credibility models to some extent. The most similar work to ours is using human rationales to improve neural predictions~\cite{bao2018deriving}. However, their work is exclusively designed to regularize the intrinsically interpretable model, i.e., attention model. In contrast, our method is widely applicable to different network architectures, including both interpretable models and black-box models, such as CNN and LSTM.

\section{Problem Statement}
In this section, we first introduce the basic notations used in this paper. Then we present the problem of learning credible deep neural network models.

\vspace{2pt}
\noindent\textbf{Notations}: Consider a typical multi-class text classification task.  
Given a training dataset which consists of $N$ instances: $D =  \{(x_1,y_1),...(x_N,y_N)\}$.
Each input text $x_n$ is composed of a sequence of $T$ words: $x_n = \{x_n^{(1)},..., x_n^{(T)} \}$, where $x_n^{(t)} \in \mathbb{R}^d$ denotes the embedding representation of the $t$-th word. Each $y_n \in \{1,2,...,C\}$ belongs to one of the $C$ output classes. Part of the training data, with a number of $N_r$, contains not only input-label pairs $(x_n,y_n)$, but also rationale $r_n$ from domain expert, with two illustrate examples shown in Fig.~\ref{fig:rationale-example}. Each entry of the expert rationale $r_n^{(t)} \in \{0,1\}$, where $1$ indicate that word $x_n^{(t)}$ is actually responsible for the prediction task, and vice versa. 

\vspace{3pt}
\noindent\textbf{Learning Credible DNNs}: The goal is to learn a DNN-based classification model which maps a text input $x_n$ to the probability output $f(x_n)$. 
We expect a trained DNN to rely on correct evidences to make decisions and pay more attention to words within the rationales. That is, for a trained DNN, the generated local explanation for each testing instance should align well with expert rationales.

\vspace{6pt}
\section{Proposed CREX Framework}
In this section, we introduce the CREX framework, which aims to regularize the local explanation when training a DNN for the task of interest, so as to promote its credibility and generalization. Besides feeding labels as supervised signals, we also enforce the explanations of the DNN predictions to conform with expert rationales, and encourage the explanations to be sparse if rationales are absent. In this way, the trained network could make predictions based on the correct evidences that we expect it to focus on.

\subsection{Augmenting Local Explanation} 

The general idea of DNN local explanation is to attribute the prediction of a DNN to its input, producing a heatmap indicating the contribution of each feature in the input to the prediction. 
There are several key desiderata for the augmented local explanation method in this work:
\begin{itemize}[leftmargin=*]
\setlength\itemsep{0em}
\item \emph{Faithful}: The provided explanations should be of high fidelity with respect to predictions of the original model.
\item \emph{Differentiable}: 
We expect the explanation method to be end-to-end differentiable, amenable for training with back-propagation and updating DNN parameters.

\item \emph{Model-agnostic}:
It is desirable that the explanation method to be agnostic to network architectures, and thus generally applicable to different networks, e.g., CNNs and LSTMs.
\end{itemize}

The explanation of prediction $f(x_n)$ for input $x_n$ is a matrix $s_n \in \mathbb{R}^ {T \times C}$, where $s_n^{(t,c)}$ denotes the contribution of word $x_n^{(t)}$ towards prediction $f_c(x_n)$ for output class $c$. We utilize an omission based method~\cite{li2016understanding} to measure the contribution of $x_n^{(t)}$, denoted as below:
\begin{equation}
s_n^{(t,c)} =  f_c(x_n) - f_c(x_n^{(\setminus t)}),
\label{equ:explanation}
\end{equation}
which quantifies the deviation of the prediction between the original input $x_n$ and the partial input $x_n^{(\setminus t)} = x_n^{(1:t-1)} \oplus x_n^{(t+1:T)}$ with $x_n^{(t)}$ omitted.
The motivation is that more important features, once being changed, will cause more significant variation to the prediction score.
It is worth noting that the omission operation may lead to invalid input, which could trigger the adversarial side of DNNs. To reflect model behaviors under normal conditions, phrase omission is conducted instead of individual word omission. Formally, we compute the contribution of $x_n^{(t)}$ by averaging the prediction changes of deleting different length-$m$ phrases that contain $x_n^{(t)}$:
\begin{equation}
s_n^{(t,c)} = \frac{1}{m} \sum_{j=1}^m [f_c(x_n) - f_c(x_n^{(1:t-1-m+j)} \oplus x_n^{(t+j:T)})].
\label{equ:explanation2}
\end{equation}
For long text classification, such as documents, we segment each original text into sentences and sequentially perform omission for each sentence. In such scenario, sentence-level contribution scores are obtained as explanation, rather than word-level scores. Both phrase omission and sentence omission could increase the faithfulness of explanation, compared with directly removing individual words~\cite{kadar2017representation}.

\subsection{Aligning Explanations with Rationales}
The key idea of CREX is that DNNs should rely on reasonable evidences to make decisions rather than bias or artifacts.
We encourage the explanation to align well with expert rationales when they are available, by considering two complementary conditions as follows. First, for the original input, we encourage the generated explanation to be \emph{confident} and focus on the relevant features as indicated by rationales. Second, for the negative input, where the important features are suppressed, the explanation should be \emph{uncertain} and have relatively uniform contribution across classes. 

\vspace{3pt}
\subsubsection{Confident Explanation}
We first feed original input $x_n$ to DNN and get model output $f(x_n)$ and explanation $s_n$.
The rationale $r_n$ points out which subset of features is important and the rest to be irrelevant. Intuitively, we achieve credibility by encouraging dense contribution scores on known important factors and encouraging sparse contribution scores on the remaining irrelevant features. We define a \emph{confident explanation} loss ($g_{conf}$), which encourages the explanation to concentrate on rationales:
\begin{equation}
g_{conf} (x_n) =\frac{1}{C} \sum_{c=1}^{C} ||(1-r_n)\odot  s_n^{(:,c)})||_1.
\label{equ:credibilityfunction}
\end{equation}
The loss aims to shrink the contribution scores of irrelevant features, in order to discourage models from capturing training data specific biases. An implicit effect of this loss is to encourage $f$ to give dense explanation scores to the relevant features, thus making $f$ pay more attention to them. As a result, the final explanation scores tend to aligning well with rationales. In addition, we observe that summing all categories $\{1,...C\}$ could yield better results comparing to only using label $y_n$ when imposing confident explanation regularization to instance $x_n$.

\begin{figure}
  \centering
  \includegraphics[width=0.87\linewidth]{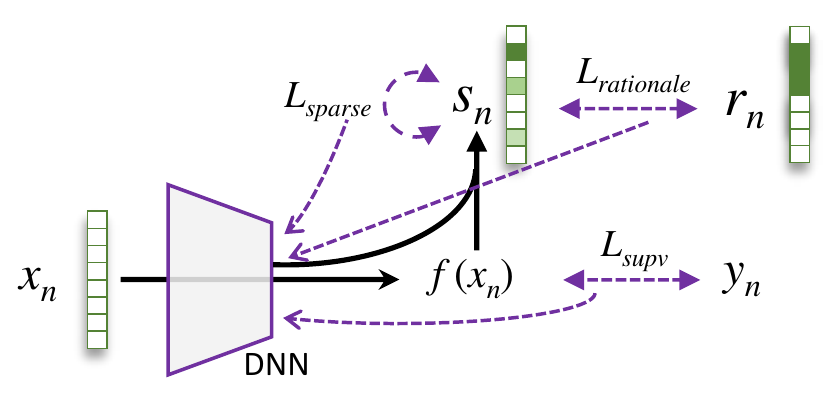}
  \vspace{6pt}
  \caption{Schematic of CREX. Black solid lines denote forward pass. Dashed line with arrows on both ends are losses. Dashed line with arrows on one side denote flow of gradients. Three vectors from left to right are input, explanation and rationale, respectively. CREX is DNN architecture agnostic, end-to-end trainable, and simple to implement.}
  \label{fig:pipeline}
\end{figure}

\vspace{3pt}
\subsubsection{Uncertain Explanation}
When the subset of important features, as indicated in $r_n$, is deleted in the original input $x_n$, we expect the DNN model to become uncertain about which category to output. This kind of inputs, named as negative inputs, are generated as the Hadamard product between the original input $x_n$ and the reversed rationale vector $(1-r_n)$:
\begin{equation}
  x^{\prime}_n = x_n \odot (1-r_n).
\end{equation}
For instance, the negative input corresponding to the first input in Fig.~\ref{fig:rationale-example} is \emph{``The movie is that even the most casual viewer may notice the''}. The intuition is that after feeding the negative input $x^{\prime}_n$ to a DNN model, we expect its probability output for ground truth label $y_n$ to be much smaller comparing to the probability value of original input $x_n$, since $x^{\prime}_n$ lacks the evidence supporting the prediction. At the same time, the contributions of different words/sentences should be distributed uniformly. Its implicit effect is to encourage the DNN model to give lower explanation scores to the features not belonging to rationales.
We first calculate the absolute value of explanation for $x^{\prime}_n$ as $\hat{s}_n^{(:,y_n)} = |s_n^{(:,y_n)}|$, and 
then normalize it as:
\begin{equation}
  e_n^{(t,y_n)} = \hat{s}_n^{(t,y_n)} / \sum_{k=1}^T \hat{s}_n^{(k,y_n)}
\end{equation}
The resultant $e_n^{(:,y_n)}$ can be seen as the soft-attention assigned by DNN for $x^{\prime}_n$. 
 After that, we define an \emph{uncertain explanation} loss ($g_{unc}$):
\begin{equation}
  g_{unc} (x^{\prime}_n) =  -  |f_{y_n}(x_n) - f_{y_n}(x^{\prime}_n)| - \alpha\, \cos(e_n^{(:,{y_n})}, q) ,
  \label{equ:uncertain}
\end{equation}
where $q$ is the discrete uniform distribution denoted as $\mathcal{U}(1,T)$, and $\alpha$ is used to balance probability output and explanation distribution.
The cosine similarity is employed to encourage explanation scores to be distributed uniformly. 

We linearly combine the two loss functions at hand, and calculate the average value over all training instances with rationales as the \emph{explanation rationale} loss, formulated as follows:
\begin{equation}
  \mathcal{L}_{rationale} = \frac{1}{N_r} \sum_{n=1}^{N_r} [g_{conf} (x_n) + \beta g_{unc} (x^{\prime}_n)].
\end{equation}
Parameter $\beta$ is utilized to balance the confident explanation and uncertain explanation.
By encouraging confident explanations to conform with rationales in original input $x_n$, and suppressing the probability output as well as explanation values in a negative input $x^{\prime}_n$, $\mathcal{L}_{rationale}$ regulates a DNN to learn useful input representations from features belonging to rationales and omit information in the irrelevant feature subset.

\subsection{Self-guidance When Rationale not Available}
In last section, given expert rationales, we render the local explanation of each instance to conform with its rationale.
However, expert rationales may not always be available. In practice, the experts may only annotate
a small ratio of training data. This could be done either when annotating a new corpus, or when adding rationales post-hoc to an existing corpus.
To guide the DNN model to focus on correct evidences in such scenario, we enforce the generated local explanation vector to be sparse for training instances without rationales. 
Simpler explanations are more credible, otherwise the dense dependencies could make it hard to disentangle the patterns in the input that actually trigger a prediction~\cite{peters2018interpretable,malaviya2018sparse,lipton2016mythos}.
To achieve this, we propose the \emph{sparse explanation} loss for those instances without rationales, denoted as follows:
\begin{equation}
\begin{aligned}
  \mathcal{L}_{sparse} =  \frac{1}{(N-N_r) \cdot C} \sum_{n=N_r + 1}^{N} \sum_{c=1}^{C} ||s_n^{(:,c)}||_1,
\end{aligned}
\label{equ:sparse}
\end{equation}
where the $\ell_{1}$ norm helps produce sparse contribution vectors. Note that this summation is performed over the $(N$-$N_r)$ instances which have no rationales.

\setlength{\textfloatsep}{14pt}
\begin{algorithm}[t!]
\DontPrintSemicolon
\KwIn{Training data $D = \{(x_n,y_n)\}_{n=1}^N$, validation data $D_v = \{(x_n,y_n)\}_{n=1}^{N_v}$, and rationales $\{r_n)\}_{n=1}^{N_r}$.} 
 Set hyperparameters $\alpha,\beta, \lambda_1,  \lambda_2$, learning rate $\eta$, iteration number $max\_iter=10$, sample index $i \in \{1,...,n\}$, and epoch index $t=0$;\;
 Initialize DNN parameters $\textbf{W}$;\;
 \While { $t \leq \text{max\_iter} $}{
$\mathcal{L}_{supv} =  \frac{1}{N} \sum_{n=1}^{N}\sum_{c=1}^{C} - \mathbb{1}{(y_{n} =c)}  \cdot \text{log}(f_c(x_n))$;\;
$\mathcal{L}_{rationale} = \frac{1}{N_r} \sum_{n=1}^{N_r} g_{conf} (x_n) + \beta g_{unc} (x^{\prime}_n)$;\;
$\mathcal{L}_{sparse} =  \frac{1}{(N-N_r) \cdot C} \sum_{n=N_r + 1}^{N} \sum_{c=1}^{C} ||s_n^{(:,c)}||_1$;\;
$\mathcal{L}(\theta, x, y, r)=  \mathcal{L}_{supv} +\lambda_1 \mathcal{L}_{rationale} +\lambda_2 \mathcal{L}_{sparse}$\;
$\textbf{W}_{t+1} = Adam(\mathcal{L}(\theta, x, y, r),\eta)$;\;
$\text{Get DNN accuracy on validation set} \, D_v$;\;
$t=t+1$;\;
}
\KwOut{ DNN $f$ with best accuracy on validation set.}
\caption{Learning credible DNNs.}
\label{alg:CREX}
\end{algorithm}
\subsection{CREX Training} 
Besides regularizing the local explanations for DNN predictions, we also expect the DNN model to learn from the ground truth labels, which is defined using supervised cross-entropy loss function as follows:
\begin{equation}
\begin{aligned}
  \mathcal{L}_{supv} =  \frac{1}{N} \sum_{n=1}^{N}\sum_{c=1}^{C} - \mathbb{1}{(y_{n} =c)}  \cdot \text{log}(f_c(x_n)). 
\end{aligned}
\label{equ:cross-entropy}
\end{equation}
Our final model is learned by balancing the supervised approximation to the labels and the conformation to expert rationales. We propose the training objective by jointly minimizing the losses as below:
\begin{equation}
\begin{aligned}
  \mathcal{L}(\theta, x, y, r)=  \mathcal{L}_{supv} +\lambda_1 \mathcal{L}_{rationale} +\lambda_2 \mathcal{L}_{sparse}.
\end{aligned}
\end{equation}
Parameters $\lambda_1$ and $\lambda_2$ are utilized to balance the supervised loss, rationale loss and sparse loss. For those $N_r$ inputs coupled with expert rationales, we impose rationale loss, while for the rest $N-N_r$ inputs we regularize them with sparse loss.
The overall idea of CREX is illustrated in Fig.~\ref{fig:pipeline}, and the learning algorithm of CREX is presented in Algorithm~\ref{alg:CREX}. Our framework is designed to train the DNN model which could make highly accurate predictions (the first term) as well as make decisions by relying on the correct evidences (the last two terms). 
In addition, our CREX training framework can be treated as knowledge distillation process that transfers expert knowledge from rationales to DNN parameters in order to yield more credible models. 
CREX is also general, and can be added to any DNN models, e.g., CNNs and LSTMs, in order to enhance model's credibility.

\section{Experiments}
In this section, we evaluate the proposed CREX framework on several real-world datasets and present experimental results in order to answer the following four research questions. 

\begin{itemize}[leftmargin=*]
\item \textbf{RQ1} - Does CREX enhance the credibility of DNNs by regularizing the local explanation using expert rationales in the training process?
\item \textbf{RQ2} - Does CREX promote the generalization of DNNs when processing unseen instances, especially for those data beyond test set?
\item \textbf{RQ3} - How do CREX components and hyperparameters affect DNNs' performance?
\item \textbf{RQ4} - How do the quantity and quality of expert rationales influence the performance of DNNs trained by CREX?
\end{itemize}

\subsection{Experimental Setup}
In this section, we introduce the overall setup of the experiments, including: \RNum{1}. DNN architectures, \RNum{2}. datasets, \RNum{3}. baseline methods, and \RNum{4}. implementation details.

\vspace{3pt}
\subsubsection{DNN Architectures}
We consider three representative DNN architectures for text classification, including CNN~\cite{kim2014convolutional}, LSTM~\cite{hochreiter1997long}, and Self-attention model~\cite{lin2017structured}. 

\vspace{3pt}
\noindent\textbf{CNN}: This is a 2-D convolutional network. The convolution operation is performed on embedding input $\{x_n^{(1)},..., x_n^{(T)} \}$ using three sizes of kernel: [2, 3, 4]. We will use ReLU activation after the convolution operation and then apply max pooling operation for every channel. Finally, the resulting tensors will be concatenated as final input representation. 

\vspace{2pt}
\noindent\textbf{LSTM}:  After feeding the input $x_n = \{x_n^{(1)},..., x_n^{(T)} \}$ to the LSTM model, $T$ hidden state vectors $\{h_n^{(1)},..., h_n^{(T)} \}$ are obtained. The dimension of each hidden state vector is 150. Max pooling is performed after all $T$ hidden vectors to obtain the final input representation. 

\vspace{2pt}
\noindent\textbf{Self-attention}: A bidirectional LSTM is first utilized to learn input representations with hidden size of 300. Then the self-attention mechanism is applied on top of LSTM representations to produce a matrix embedding of the input sentence. This matrix contains 10 embeddings, where every embedding represents an encoding of the input sentence but giving an attention to a specific part of the sentence. These embeddings are concatenated as the final input representation.

\noindent For all three networks, after transforming variable length sentences into fixed size representations, fully connected layers are added after the representations to get logits~\cite{hinton2015distilling} for multiple output classes. Finally, a softmax layer is added to convert logits to probability outputs.

\begin{table}
\renewcommand{\arraystretch}{1.2}
\setlength{\tabcolsep}{3pt}
\small
\centering
\begin{tabular}{r c c c c }
\toprule
Dataset &Train & Dev & Test & \small{Text length}  \\
\midrule
Movie Review (MR)   &1,500 & 100 & 200 & 794  \\
Product Review (PR)   &4,000  & 473  & 1,700 & 113  \\
\bottomrule
\end{tabular}
\vspace{3pt}
\caption{Dataset statistics of MR and PR dataset, including number for training, development and test set, as well as average text length.}
\label{fig:datasetStat}
\end{table}

\vspace{4pt}
\subsubsection{Datasets and Rationales}
We consider two benchmark text classification datasets. Both datasets are randomly split into training, development and test set, the statistics of which are reported in Tab.~\ref{fig:datasetStat}.

\vspace{2pt}
\noindent\textbf{Movie Review Dataset (MR)}: It is a binary sentiment classification dataset with movie reviews from IMDB~\cite{pang2004sentimental}. Originally, this dataset is obtained by crawling movie reviews from the Internet Movie Database (IMDB), consisting 1000 positive and 1000 negative movie reviews~\cite{pang2004sentimental}. 
Zaidan et al.~\cite{zaidan2007using} supplemented this dataset rationales for 1800 documents~\footnote{\url{http://www.cs.jhu.edu/~ozaidan/rationales/}}.
The rationales used in this dataset are sub-sentential snippets with a higher relevance for prediction task~\footnote{In terms of the rationale collection process, the agreement among different annotators, as well as the time complexity of rationale annotations, we refer interested readers to the work by Zaidan et al.~\cite{zaidan2007using}.}, with illustrative example shown in Fig.~\ref{fig:rationale-example}. The average length per rationale for per input text is 125, while the average text length is 794. Comparing to the whole text, the rationale is sparse. 

\vspace{2pt}
\noindent\textbf{Product Review Dataset (PR)}: 
It is a multi-aspect beer review dataset~\cite{mcauley2012learning} with data derived from BeerAdvocate~\footnote{\url{https://www.beeradvocate.com/}}. This dataset contains reviews for three aspects of beer: appearance, aroma and palate, where we only distinguish appearance. Originally this dataset contains reviews with rating in the range of $[0,1]$. Similar to ~\cite{bao2018deriving}, we consider this as binary classification task, by labelling ratings $\leq$0.4 as negative category, while labeling those $\geq$0.6 as positive category. Rationales are provided by~\cite{lei2016rationalizing}, which are also sub-sentential snippets indicating higher relevance for prediction (see Fig.~\ref{fig:rationale-example}). The rationale within this dataset is also sparse, with an average length of 19, comparing to average text length of 113.

\vspace{4pt}
\subsubsection{Baseline Methods}
We evaluate effectiveness of CREX by comparing it with three baseline approaches. 
\begin{itemize}[leftmargin=*]
\setlength\itemsep{0em}
\item \emph{Vanilla DNN}: This is the most typical way to train DNN for text classification tasks. DNN models are trained with only standard cross entropy loss, optimizing parameters to minimize Eq. (\ref{equ:cross-entropy}). 

\item \emph{Data Augmentation}: Back translation is an effective data augmentation method to boost model performance, e.g., machine translation~\cite{sennrich2015improving,poncelas2018investigating}. The original text is first translated to an intermediate language (we use German) and then translated back to English via the Google Translate API~\footnote{\url{https://pypi.org/project/googletrans}}. The motivation is to use synonym replacement and sentence paraphrase to avoid overfitting to functional words. 
\item \emph{Rationale Augmentation}: Expert rationales are extracted from the original text as additional training instances. These data are incorporated with original training data, resulting a final training dataset of double size comparing with original one. The intuition is to explicitly push DNNs to focus on rationales to make decisions.
\end{itemize}

\begin{table}
\renewcommand{\arraystretch}{1.2}
\setlength{\tabcolsep}{3pt}
\centering
\small
\begin{tabular}{l c c c c c c}
\toprule 
& \multicolumn{3}{c}{MR} & \multicolumn{3}{c}{PR} \\
\cmidrule(l){2-4} \cmidrule(l){5-7}
Models & CNN & LSTM & Atten  & CNN & LSTM & Atten\\ 
\midrule 
Vanilla DNN & 2.86 & 2.67 & 2.40  & 3.96 & 3.77 & 3.73\\ 
Data Augment & 2.75 & 3.20 & 2.29  & 3.85 & 3.70 & 4.16\\ 
Rationale Augment & 2.52 & 2.45 & 2.25 & 3.65 & 3.61 & 3.59\\ 
CREX & \textbf{2.24} & \textbf{2.38} & \textbf{1.91}  & \textbf{3.52} & \textbf{3.54} & \textbf{3.15}\\
\midrule
Parameter $\lambda_1$ & 5e-2 & 1e-3 & 2e-4  & 1e-4 & 2e-4 & 1e-4\\ 
Parameter $\alpha$ & 0.2 & 0.5 & 0.3 & 0.5 & 0.3 & 0.5\\ 
\bottomrule 
\end{tabular}
\vspace{4pt}
\caption{Credibility statistical comparisons of three DNN architectures on MR and PR test set, and corresponding optimal hyperparameter settings.}
\label{tab:credibilityEvaluation}
\end{table}

\vspace{4pt}
\subsubsection{Implementation Details}
We use the pre-trained 300-dimensional word2vec~\footnote{\url{https://code.google.com/archive/p/word2vec/}} word embedding~\cite{mikolov2013distributed} to initialize the embedding layer for all three architectures. For those words that do not exist in word2vec, their embedding vectors are initialized  with some random values. We tune the learning rate over the range $\{1e$-4, $1e$-3, $1e$-2, $1e$-1$\}$ and utilize Adam optimizer~\cite{kingma2014adam} to optimize these models. For each model, all hyperparameters are tuned using the development set, according to the accuracy and credibility performance. Optimal values of $\alpha$ and $\lambda_1$ for different models are listed in Tab.~\ref{tab:credibilityEvaluation}, while $\beta$ and $\lambda_2$ are fixed as 1 and $1e$-5 respectively for all models. To avoid overfitting, we apply dropout to fully connected layers for all DNN models~\cite{srivastava2014dropout}. We implement all DNN models using the PyTorch library. Each model is trained for ten epoches and the one with the best performance on the development set is selected as the final model. In our experiments, all DNN models could converge within 10 epoches, and increasing the number may lead to overfitting. Besides, since all models use random initialization, which leads to variance in performances at different runs. Therefore, we report the average values over three runs for all DNNs in the following experiments.


\subsection{Credibility and Accuracy on Test Set}\label{credibilityAndAccuracy}
In this section, we evaluate the performance of all trained DNNs on test set. Two metrics are employed for evaluation: credibility and prediction accuracy. The credibility here is defined as the extent of agreement between the generated DNN local explanations and expert rationales. 

\begin{table}
\renewcommand{\arraystretch}{1.2}
\setlength{\tabcolsep}{3.2pt}
\centering
\small
\begin{tabular}{l c c c c c c}
\toprule 
& \multicolumn{3}{c}{MR} & \multicolumn{3}{c}{PR} \\
\cmidrule(l){2-4} \cmidrule(l){5-7}
Models & CNN & LSTM & Atten  & CNN & LSTM & Atten\\ 
\midrule 
Vanilla DNN & 93.7 & 93.2 & \textbf{94.7}  & \textbf{94.9} & 94.5 & 94.3\\ 
Data Augment & 91.0 & 88.3 & 90.1  & 94.7 & 94.5 & 93.9\\ 
Rationale Augment & \textbf{94.0} & 94.2 & 93.8 & 94.3 & 95.1 & 94.1\\ 
CREX & 93.8 & \textbf{94.3} & 94.5  & 94.2 & \textbf{94.8} & \textbf{94.5}\\
\bottomrule 
\end{tabular}
\vspace{4pt}
\caption{Accuracy comparisons (in percent) of CREX and baseline methods for three DNN architectures on MR and PR test set.}
\label{tab:Accuracy-test-set}
\end{table}

\vspace{3pt}
\subsubsection{Quantitative Evaluation of Credibility}
To measure credibility, we calculate the matching degree between local explanation of DNN prediction with rationale. Specifically, We use the symmetric KL divergence between the normalized absolute value of explanation $s_n$ and the normalized rationale $r_n$: 
\begin{equation}
symKL(s^{\prime}_n, r^{\prime}_n) = \frac{1}{2}[KL(s^{\prime}_n || r^{\prime}_n) + KL(r^{\prime}_n || s^{\prime}_n)]
\end{equation}
where lower divergence means higher credibility~\cite{wang2018learning}.
We compare the credibility scores of CREX with three baseline methods on three DNN architectures over MR and
PR dataset. The credibility results are presented in Tab.~\ref{tab:credibilityEvaluation}. Comparing with Vanilla, the relative improvement of CREX is encouraging, with KL divergence drops ranging from 0.29 to 0.62 for DNNs in MR, from 0.23 to 0.58 for DNNs in PR. This ascertains the effectiveness of CREX in boosting the credibility of DNNs by pushing them to employ correct evidences to make decisions. The increased credibility of Rationale Augmentation comparing to Vanilla DNN also validates the value of expert knowledge, which succeeds to push models to focus more on evidences in the rationales to make decisions. In contrast, using back translation as Data Augmentation cannot always enhance the model credibility.

\vspace{3pt}
\subsubsection{Quantitative Evaluation of Accuracy}
DNNs trained via CREX have comparable predictive accuracy with the three baselines on MR and PR test set, as shown in Tab.~\ref{tab:Accuracy-test-set}. Besides, the results of three comparing methods, including Vanilla training, Rationale augmentation, and CREX, are not substantially different. It means that the increased credibility does not sacrifice model performance on test set.

\begin{figure}
  \centering
  \includegraphics[width=1\linewidth]{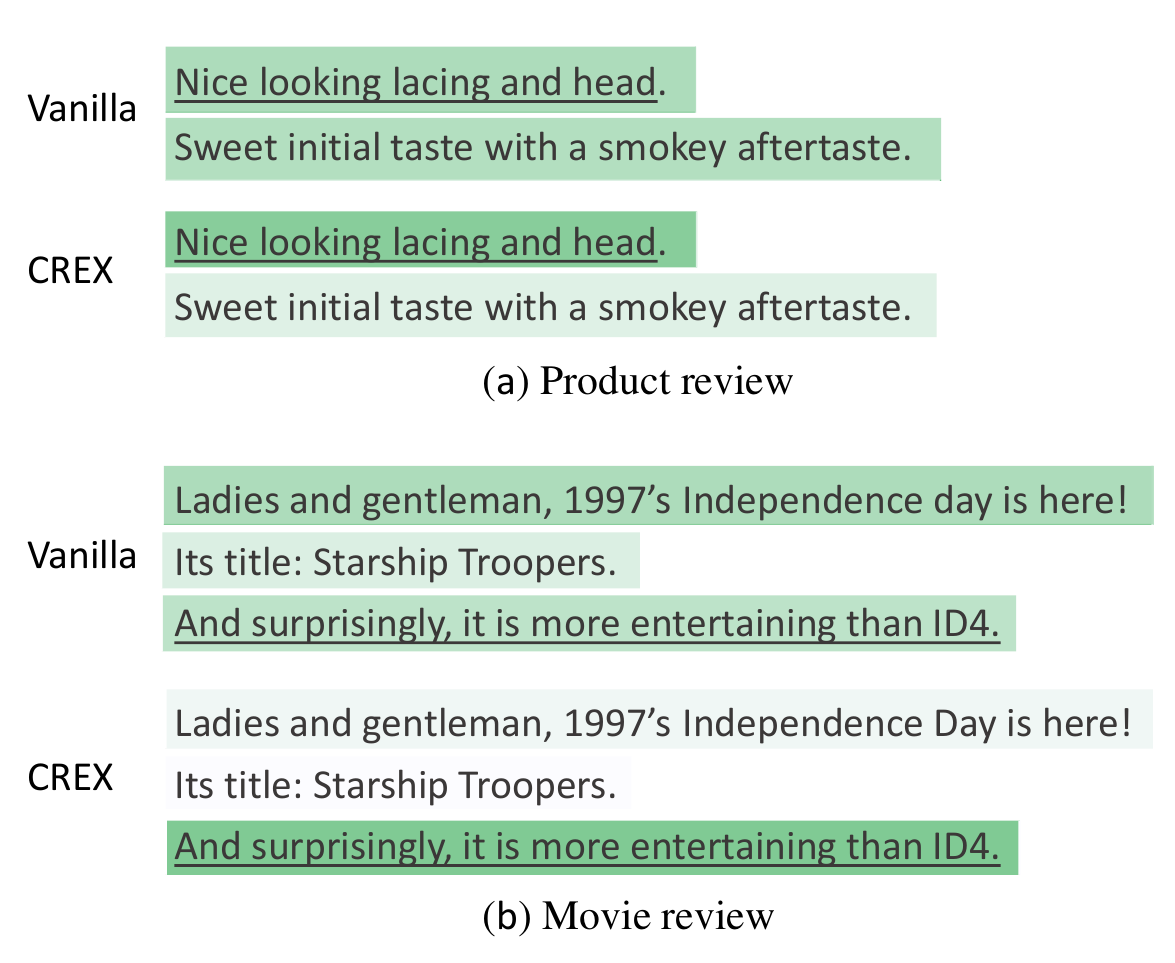}
  \caption{Sentence-level explanation heatmap comparison between CREX and Vanilla DNN. Ground truth is annotated with underline. (a) Beer appearance review, positive label. (b) Movie review, negative sentiment label. Here ID4 denotes the movie Independence Day. }
  \label{fig:heatmap}
\end{figure}

\vspace{3pt}
\subsubsection{Qualitative Evaluation of Credibility}\label{casestudy}
We provide case studies to qualitatively show the effectiveness of the increased credibility, as shown in Fig.~\ref{fig:heatmap}. We show the sentence-level explanation scores, where deeper color means higher contribution to the prediction. For both cases, these two predictions are made by self-attention model, trained via Vanilla method and CREX method respectively.

For the first product review (PR) case shown in Fig.~\ref{fig:heatmap} (a), both DNNs give positive prediction for this testing instance, with 99.9\% and 99.7\% confidence respectively. 
We can observe that the Vanilla DNN pays nearly equal attention to the second sentence as the first one, even though the second sentence talks about the beer palate (\emph{``sweet'', ``taste'', ``aftertaste''}) and has nothing to do with beer appearance. It indicates that the DNN classifier may have overfitted to bias in training set. In contrast, CREX could push the DNN to rely on correct evidences relevant to beer appearance, i.e., \emph{``good looking''}, to make decisions. This explanation is consistent with our human cognition, and thus CREX is more likely to earn trust from end-users. 

Similarly for the movie review case in Fig.~\ref{fig:heatmap} (b), although both self-attention models give correct predictions, they use distinct evidences to make decisions. Vanilla DNN pays nearly equal attention to the first and third sentence, where only the third sentence contains more generalizable features. One possible reason to explain this phenomenon is that the DNN may have memorized movie-unique terms to make decisions, which is supposed to perform poorly in movie reviews beyond training and test data. In contrast, CREX could focus mostly on the third sentence with task-relevant adjective i.e., \emph{``entertaining''}, to make positive sentiment prediction. This finding demonstrates that CREX is able to disentangle useful knowledge from dataset specific biases. In next section, we will demonstrate the benefit of increased credibility of CREX on unseen testing data which are not drawn from test set.

\subsection{Generalization Accuracy beyond Test Set}\label{beyondtestet}
Currently, the generalization performance of DNNs is usually calculated using the prediction accuracy on the held-out test set. This is problematic due to the independent and identically distributed (i.i.d.) training-test split of data, especially in the presence of strong priors~\cite{agrawal2018don}. The DNN model can succeed by simply recognize patterns that only happen to be predictive on instances over the test set~\cite{minervini2018adversarially}. As evidenced by the example in Sec.~\ref{casestudy}, the DNN may rely on the aroma and palate as evidences to support appearance prediction, which is supposed to perform poorly in beer reviews outside of the training and test data. Consequently, test set fails to adequately measure how well DNN systems perform on new and previously unseen inputs. To assess the true generalization ability of DNN models as well as to demonstrate the benefit of increased credibility of CREX, we also evaluate the model performance using data beyond the test set. 

\begin{table}
\renewcommand{\arraystretch}{1.2}
\setlength{\tabcolsep}{3.0pt}
\centering
\small
\begin{tabular}{l c c c c c c}
\toprule 
& \multicolumn{3}{c}{Kaggle} & \multicolumn{3}{c}{Polarity} \\
\cmidrule(l){2-4} \cmidrule(l){5-7}
Models & CNN & LSTM & Atten  & CNN & LSTM & Atten\\ 
\midrule 
Vanilla DNN & 74.3 & 73.6 & 74.7  &60.7 & 62.6 & 64.8\\ 
Data Augment & 75.7 & 70.3 & 75.0  & 62.5 & 58.1 & 65.4\\ 
Rationale Augment & 76.5 & 73.9 & \textbf{75.8} & 63.1 & 63.2 & 65.3\\ 
CREX & \textbf{78.4} & \textbf{75.7} & 75.2  & \textbf{63.2} & \textbf{63.8} & \textbf{65.7}\\
\bottomrule 
\end{tabular}
\vspace{4pt}
\caption{Generalization accuracy (in percent) of DNNs trained using MR dataset on two alternative datasets: Kaggle and Polarity.}
\label{tab:generalizationAccuracyMR}
\vspace{5pt}
\end{table}

\begin{table}
\renewcommand{\arraystretch}{1.2}
\setlength{\tabcolsep}{6pt}
\centering
\small
\begin{tabular}{l c c c}
\toprule 
Models & CNN & LSTM & Atten  \\ 
\midrule 
Vanilla DNN & 92.1 & 91.5 & 91.0  \\ 
Data Augment & 92.4 & 92.1 & 90.1  \\ 
Rationale Augment & 92.5 & 91.9 & 90.9 \\ 
CREX & \textbf{92.7} & \textbf{92.3} & \textbf{91.2}  \\
\bottomrule 
\end{tabular}
\vspace{4pt}
\caption{Generalization accuracy (in percent) of DNNs trained using PR dataset on an adversarial dataset.} 
\label{tab:generalizationAccuracyPR}
\end{table}

\vspace{3pt}
\subsubsection{Generalization for DNNs Trained on MR}
For DNNs trained on MR, we use two alternative datasets: 

\begin{itemize}[leftmargin=*]
\item \textbf{Kaggle movie reviews dataset}~\footnote{\url{https://www.kaggle.com/iarunava/imdb-movie-reviews-dataset}} (\textbf{Kaggle}) It is a binary sentiment classification benchmark, with movie reviews from IMDB, consisting of 50,000 reviews.
\item \textbf{Sentence polarity dataset}~\footnote{\url{http://www.cs.cornell.edu/people/pabo/movie-review-data/}} (\textbf{Polarity})~\cite{pang2005seeing}. Another binary sentiment classification dataset with data from IMDB, consisting of 10,662 reviews.
\end{itemize}
Note that none of the data from these two datasets is utilized to train DNN models or tune hyperparameters. They only serve the testing purpose.
The generalization accuracy statistics are shown in Tab.~\ref{tab:generalizationAccuracyMR}. There are several key observations. Firstly, comparing with the accuracy in Tab.~\ref{tab:Accuracy-test-set}, there is a significant \emph{generalization gap} between predictive accuracy on MR test set and Kaggle (or Polarity), for all three architectures. Almost most of the accuracy scores are above 90\% on the corresponding test set. In contrast, all accuracy scores are below 80\% for Kaggle and below 70\% for Polarity dataset. Secondly, CREX could reduce this \emph{generalization gap} comparing to baseline methods. In Tab.~\ref{tab:generalizationAccuracyMR}, CREX DNNs achieve substantial accuracy enhancements comparing to Vanilla DNNs, with relative accuracy improvement of 4.1\%, 2.1\%, 0.5\% for three networks on Kaggle, and 2.5\%, 1.2\%, 0.9\% for three networks on Polarity. These enhancements have validated the benefit of the increased credibility of our trained DNNs. Thirdly, an interesting observation is that there exists a positive correlation between the degree of credibility and the generalization accuracy on data not existing in test set. Rationale Augmentation has consistent accuracy improvement comparing with Vanilla, while Data Augment via back translation does not, as shown in Tab.~\ref{tab:generalizationAccuracyMR}. This conforms very well with the credibility performance in Tab.~\ref{tab:credibilityEvaluation}.

\vspace{3pt}
\subsubsection{Generalization for DNNs Trained on PR}
To test generalization performance of DNNs trained on PR, we create an adversarial dataset by removing sentences relevant to beer aroma and palate. This is achieved via detecting sentences containing word \emph{``taste'', ``smell'', ``aroma'', ``flavor'', ``drinking''} from the original PR test set. Note that we only differentiate beer appearance, thus description words about beer aroma and palate are considered as training set specific bias. The corresponding accuracy is shown in Tab.~\ref{tab:generalizationAccuracyPR}, where CREX consistently outperforms baseline methods. Particularly, CREX DNNs have promoted the accuracy ranging from 0.2\% to 0.8\% comparing to Vanilla DNNs. It demonstrates that our trained DNNs rely more on correct evidences relevant to beer appearance rather than aroma and palate to make decisions, thus could achieve better generalization accuracy.

\begin{table}
\renewcommand{\arraystretch}{1.25}
\setlength{\tabcolsep}{5pt}
\centering
\small
\begin{tabular}{l c c c}
\toprule 
Models & Credibility & Kaggle & Polarity  \\ 
\midrule 
CREX\_{conf}  & 2.27 & 76.7 &  63.0 \\ 
CREX\_{unc}  & 2.37 & 77.6 &  62.2 \\ 
CREX  & 2.24 & 78.4 & 63.2  \\ 
\bottomrule 
\end{tabular}
\vspace{4pt}
\caption{Ablation analysis of CNN trained on MR dataset. The first column is credibility score on MR test set, the last two columns denote generalization accuracy on two alternative datasets.} 
\label{tab:ablation}
\vspace{4pt}
\end{table}
\subsection{Ablation Study and Hyperparameters Analysis}\label{ablations}
In this section, we utilize CNN trained on MR dataset to conduct ablation and hyperparameter analysis to study the impacts and contributions of different components of CREX.  

\begin{figure}
  \centering
  \begin{overpic}[width=1.01\linewidth]{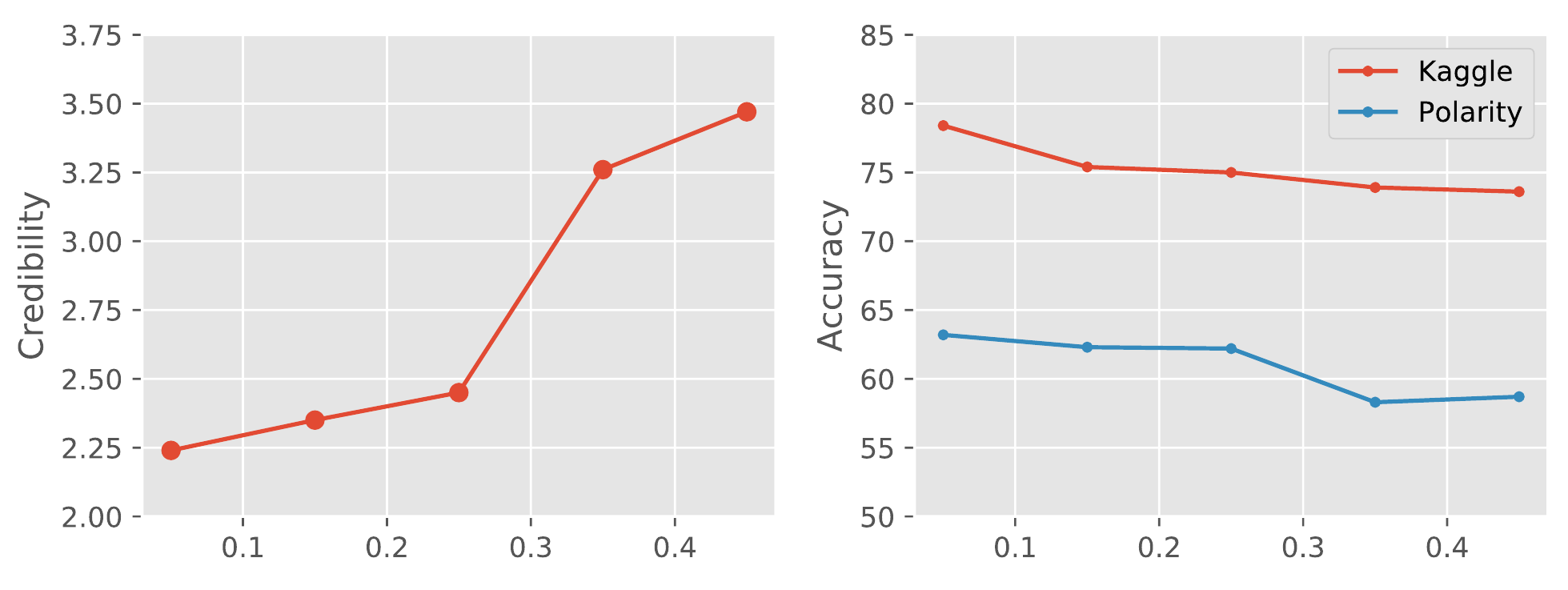}
    \put(20.0,  -3){\footnotesize (a) Credibility}
    \put(68,  -3){\footnotesize (b) Accuracy}
  \end{overpic}
  \vspace{1pt}
  \caption{CNN performance under different values of parameter $\lambda_1$. (a) credibility performance on MR test set. (b) generalization accuracy (in percent) on two alternative datasets.}
  \label{fig:hyperparameter}
\end{figure}

\vspace{3pt}
\subsubsection{Ablation Study}
We compare CREX with its ablations to identify the contributions of different components. The ablations include (\RNum{1}). CREX\_conf, using only confident explanation loss in Eq. (\ref{equ:credibilityfunction}), and (\RNum{2}). CREX\_unc, using only uncertain explanation loss in Eq. (\ref{equ:uncertain}). The comparison results between CREX and its ablations are listed in Tab.~\ref{tab:ablation}. We can observe that CREX outperforms the two ablations in terms of credibility as well as generalization accuracy on Kaggle and Polarity dataset. It indicates that these two components are complementary to each other in general, thus both are crucial in promoting model performance.

\vspace{3pt}
\subsubsection{Hyperparameter Analysis}
We evaluate the effect of different degrees of rationale loss regularization towards models' performance, by altering the value of the hyperparameter $\lambda_1$. As shown in Tab.~\ref{tab:credibilityEvaluation}, the optimal $\lambda_1$ for CNN trained on MR dataset is 5e-2. We are interested in how the model performance changes as we keep increasing the value of $\lambda_1$.  The credibility and generalization accuracy are shown in Fig.~\ref{fig:hyperparameter}.
As the value of $\lambda_1$ increases, the CNN credibility begin to drop, i.e., KL divergence increases, and the model generalization accuracy on Kaggle and Polarity also decreases. Particularly, we observe a dramatic change of credibility and accuracy when $\lambda_1$ is larger than 0.25. This indicates that the models have overfitted to rationales, which also could sacrifice generalization performance.

\subsection{Rationale Quantity and Quality Analysis}
When incorporating human knowledge with DNN models, the quantity and quality of knowledge could have significant influences. In this section, we employ CNN trained on MR dataset to analyze how the performances of neural networks would be affected by different conditions of rationale.

\begin{figure}
  \centering
  \begin{overpic}[width=1.02\linewidth]{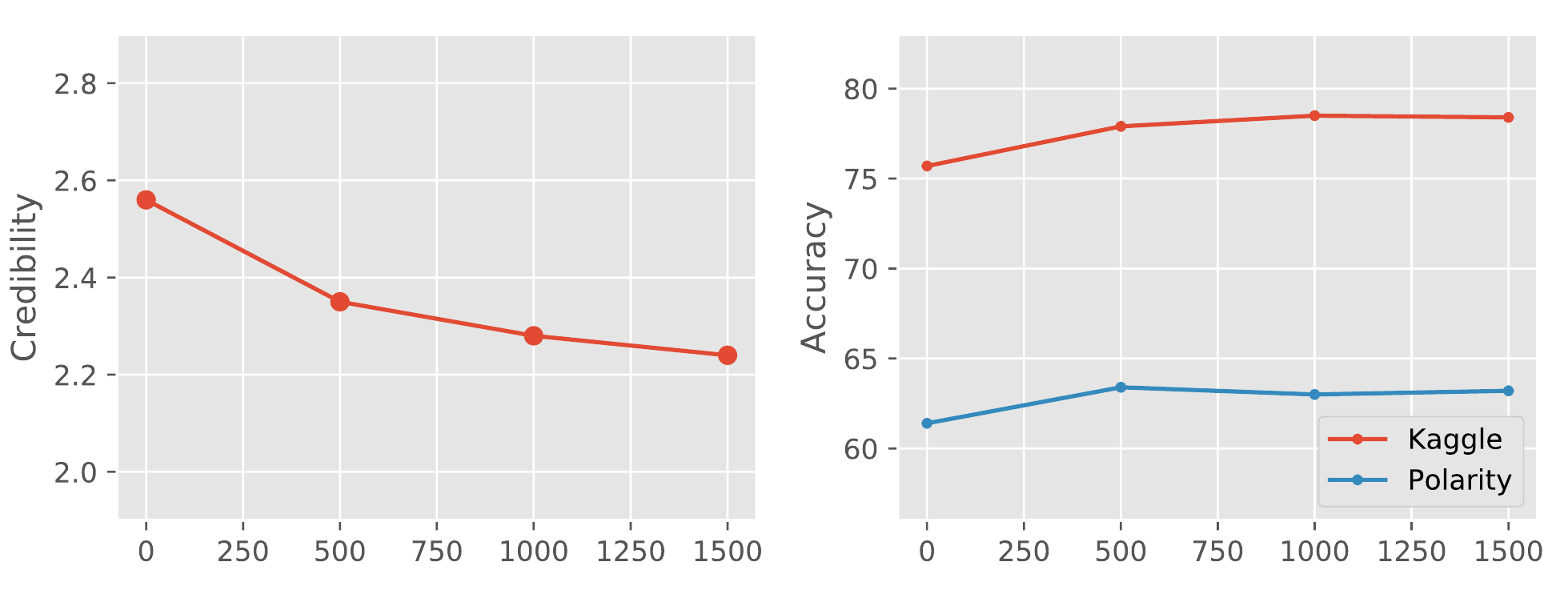}
    \put(20.0,  -3){\footnotesize (a) Credibility}
    \put(68,  -3){\footnotesize (b) Accuracy}
  \end{overpic}
  \vspace{0.5pt}
  \caption{CNN performance under different numbers of rationale. (a) credibility performance on MR test set. (b) generalization accuracy (in percent) on two alternative datasets. }
  \label{fig:numberOfRationale}
\end{figure}

\vspace{3pt}
\subsubsection{Rationale Number Analysis}
We study the effect of expert knowledge by altering the number of rationales $N_r$ in the training set, and examine the credibility and accuracy change of the trained CNN. For those instances without rationales, we impose sparse regularization as in Eq. (\ref{equ:sparse}). The results are illustrated in Fig.~\ref{fig:numberOfRationale}. There are two interesting observations. Firstly, even when rationale number $N_r$ = $0$, our CNN could achieve improved performance comparing to the Vanilla CNN. The divergence has dropped from 2.86 to 2.58 comparing to Tab.~\ref{tab:credibilityEvaluation}, and Kaggle and Polarity accuracy has increased 1.4\% and 0.7\% respectively comparing to Tab.~\ref{tab:generalizationAccuracyMR}, showing the effectiveness of sparse explanation loss in Eq. (\ref{equ:sparse}). Secondly, when the rationale number is 500, our CNN already has comparable accuracy comparing with $N_r$ = $1500$, indicating that a small ratio of rationales is sufficient for network performance promotion. Considering the annotation effort of expert rationales, this advantage of requiring small number of rationales is significant.

\vspace{3pt}
\subsubsection{Rationale Quality Analysis} 
\begin{figure}
  \centering
  \begin{overpic}[width=1\linewidth]{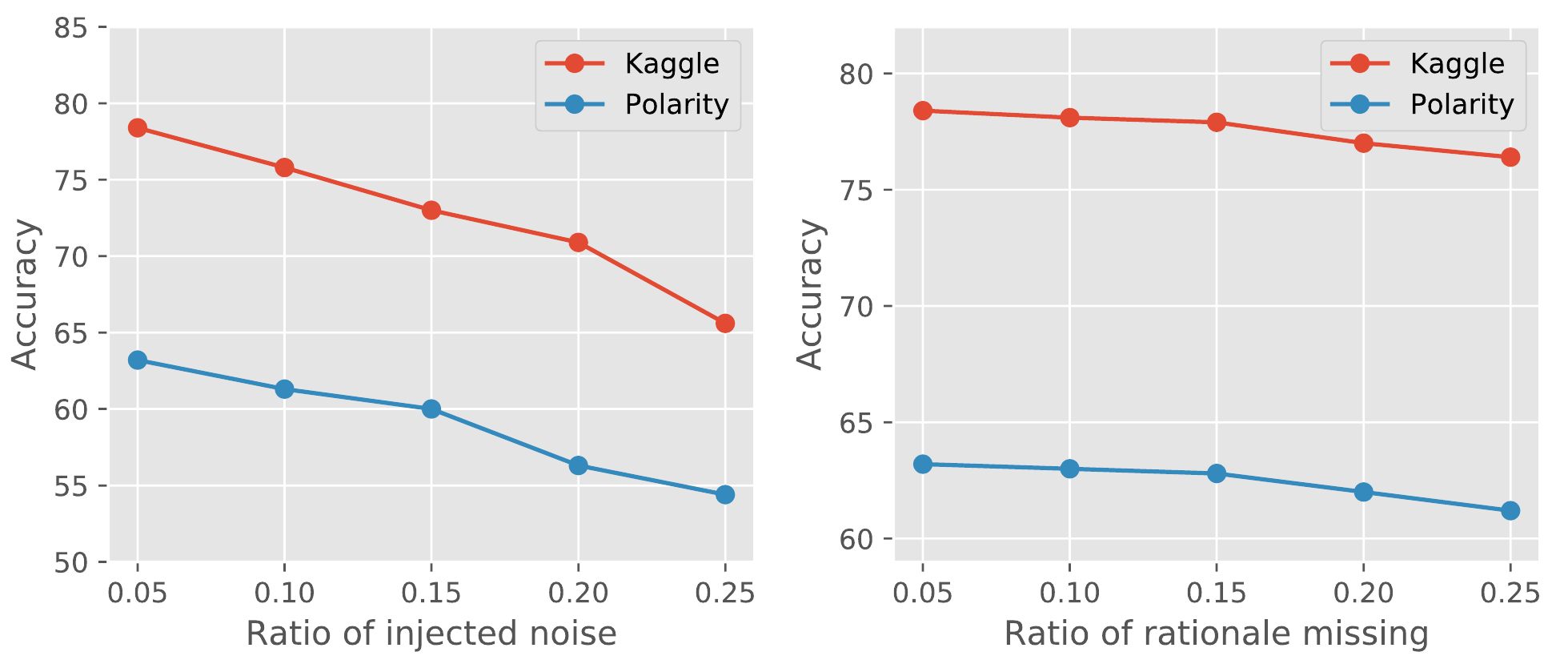}
    \put(20.0,  -3){\footnotesize (a) Mistakes}
    \put(68,  -3){\footnotesize (b) Missing}
  \end{overpic}
  \vspace{1pt}
  \caption{Rationale quality analysis using CNN generalization accuracy (in percent). (a) containing different ratios of mistakes. (b) missing different ratios of rationales. }
  \label{fig:qualityOfRationale}
\end{figure}

In this experiment, we analyze the effect of low quality rationales towards the DNN model performance. We consider two types of low quality: (\RNum{1}) containing mistakes (expert annotations could be sometimes wrong, and some irrelevant features are highlighted by the experts); (\RNum{2}) missing another set of important rationales. To simulate the first case, we inject different level of noise to the current rationales, and test model performance. Similarly, to test the second case, we delete different ratios of important features from current rationales to make the knowledge incomplete. We report CNN generalization accuracy over Kaggle and Polarity in Fig.~\ref{fig:qualityOfRationale}. There are several key findings. Firstly, the model performances are highly sensitive to rationale noise (see Fig.~\ref{fig:qualityOfRationale} (a)), where a small ratio of mistakes, e.g., 10\%, would significantly decrease generalization accuracy. Secondly, model performances are relatively robust to missing rationales (see Fig.~\ref{fig:qualityOfRationale} (b)). The reason for this phenomenon is that the remaining rationale still contains important features. By capturing sparse connections between input text and output, model could make reasonable predictions. Thirdly, considering that rationale missing is more common than containing crucial mistakes in real world rationale annotation, thus CREX is relatively robust to low-quality knowledge.

\vspace{3pt}
\subsubsection{Running Efficiency Analysis}
Due to the calculation of local explanation and regularization using rationales, the training speed of CREX is slightly slower than the Vanilla DNN training.
As shown in Tab.~\ref{tab:runningTime}, on average it takes 24 minutes to train CNN on the Movie Review dataset if using all the 1500 rationales in the training process (with our unoptimized code and using PyTorch GPU version). 
Even though CREX requires less training epoches to converge comparing to Vanilla, each epoch takes longer time than Vanilla training. To promote the training scalability of CREX, i.e., when the CREX is trained on a dataset which has much more training data comparing to MR and PR, We could reduce the ratio of rationales to speed up running of each epoch and make the total training time bearable. 
On the other hand, during the test stage, DNNs trained by CREX would need the same time (on average 8e-3 seconds) as Vanilla to yield prediction for an input, meaning that increased credibility of CREX would not sacrifice inference speed.

\begin{table}
\renewcommand{\arraystretch}{1.25}
\setlength{\tabcolsep}{3.5pt}
\centering
\small
\begin{tabular}{l c c }
\toprule 
Models & Training time  & Test time per input  \\ 
\midrule 
Vanilla CNN & 2.5 min  & 8e-3 seconds   \\ 
CREX CNN & 18.1 min  & 8e-3 seconds   \\
\bottomrule 
\end{tabular}
\vspace{5pt}
\caption{Running time comparison of Vanilla and CREX CNN. For training time, we report average value for three runs. Test time is the average over test set.} 
\label{tab:runningTime}
\end{table}

\section{Conclusion and Future Work}
There has been an increasing interest recently in developing more trustworthy DNNs.
In pursuit of this objective, we propose CREX, aiming to train credible DNNs which employ correct evidences to make decisions. We employ a specific kind of domain knowledge, called rationales, to guide the learning algorithms towards providing credible explanations, by pushing the explanation vectors to conform with rationales. CREX is DNN architecture agnostic, end-to-end trainable, and simple to implement. Experimental results show that our resulting DNN models have a higher probability to look at correct evidences rather than training dataset specific bias to make predictions. Although DNNs trained using CREX do not always improve prediction accuracy on held-out test set, they generalize much better on data which are beyond test set and which are representatives of underlying real-world tasks, highlighting the advantages of the increased credibility. 
High credibility and robustness of DNN are essential to earn trust of end-users towards a network model's predictions, and we believe the enhanced credibility and generalization will pave the way for their wider adoptions in real world. 

On the other hand, it is not guaranteed that the incorporation of human knowledge with DNN models would always promote neural network performance, unless the knowledge have sufficiently high quality.
Currently, we have explored the enhancement of DNNs via relatively high quality rationales. The low-quality knowledge issue is a challenging topic and would be explored in our future research.

\newpage

\newpage
\bibliographystyle{IEEEtran}
\bibliography{IEEEabrv,icdm}
\end{document}